\documentclass[conference]{IEEEtran}
\IEEEoverridecommandlockouts
\usepackage{cite}
\usepackage{multirow}
\usepackage{amsmath,amssymb,amsfonts}
\usepackage{algorithmic}
\usepackage{graphicx}
\usepackage{textcomp}
\usepackage{xcolor}
\usepackage{subcaption}
\usepackage{hyperref}
\usepackage[ruled,vlined]{algorithm2e}
\usepackage[symbol]{footmisc}

\hypersetup{
    colorlinks=true,
    linkcolor=red,
    filecolor=magenta,      
    urlcolor=cyan,
}
		
\newcommand{\subparagraph}{}
\usepackage{titlesec}

\makeatletter
\def\@seccntformat#1{\@ifundefined{#1@cntformat}%
  {\csname the#1\endcsname\quad}
 {\csname #1@cntformat\endcsname}}
\makeatother

\renewcommand{\thesection}{\arabic{section}.}
\renewcommand{\thesubsection}{\thesection\arabic{subsection}.}

\titleformat*{\section}{\Large\bfseries}
\titleformat*{\subsection}{\large\bfseries}
\titleformat*{\paragraph}{\large\bfseries}

\usepackage{times}

\begin{document}

\title{\textbf{Deep-CAPTCHA}: a deep learning based CAPTCHA solver for vulnerability assessment}

\author{Zahra Noury $^*$,  Mahdi Rezaei $^{\dagger}$\\
Faculty of Computer and Electrical Engineering, Qazvin Azad University\\
Faculty of Environment, Institute for Transport Studies, The University of Leeds\\
$^*$\href{mailto:zahra.noury@qiau.ac.ir}{zahra.noury@qiau.ac.ir},  $^{\dagger}$\href{mailto:m.rezaei@leeds.ac.uk}{m.rezaei@leeds.ac.uk}}

\maketitle

\begin{abstract}
\, CAPTCHA is a human-centred test to distinguish a human operator from bots, attacking programs, or other computerised agent that tries to imitate human intelligence. In this research, we investigate a way to crack visual CAPTCHA tests by an automated deep learning based solution. The goal of this research is to investigate the weaknesses and vulnerabilities of the CAPTCHA generator systems; hence, developing more robust CAPTCHAs, without taking the risks of manual try and fail efforts. We develop a Convolutional Neural Network called \textit{Deep-CAPTCHA} to achieve this goal. The proposed platform is able to investigate both numerical and alphanumerical CAPTCHAs. To train and develop an efficient model, we have generated a dataset of 500,000 CAPTCHAs to train our model. In this paper, we present our customised deep neural network model, we review the research gaps, the existing challenges, and the solutions to cope with the issues. Our network's cracking accuracy leads to a high rate of 98.94\% and 98.31\% for the numerical and the alpha-numerical test datasets, respectively. That means more works is required to develop robust CAPTCHAs, to be non-crackable against automated artificial agents. As the outcome of this research, we identify some efficient techniques to improve the security of the CAPTCHAs, based on the performance analysis conducted on the \textit{Deep-CAPTCHA} model.
\end{abstract}
%
\section{Introduction}
CAPTCHA, abbreviated for \textbf{C}ompletely \textbf{A}utomated \textbf{P}ublic \textbf{T}uring test to tell \textbf{C}omputers and \textbf{H}umans \textbf{A}part is a computer test for distinguishing between humans and robots. As a result, CAPTCHA could be used to prevent different types of cyber security treats, attacks, and penetrations towards the anonymity of web services, websites, login credentials, or even in semi-autonomous vehicles \cite{rezcvpr14} and driver assistance systems \cite{rez10a} when a real human needs to take over the control of a machine/system. 

\begin{figure}[t]
\centering
\includegraphics[width=0.85\columnwidth]{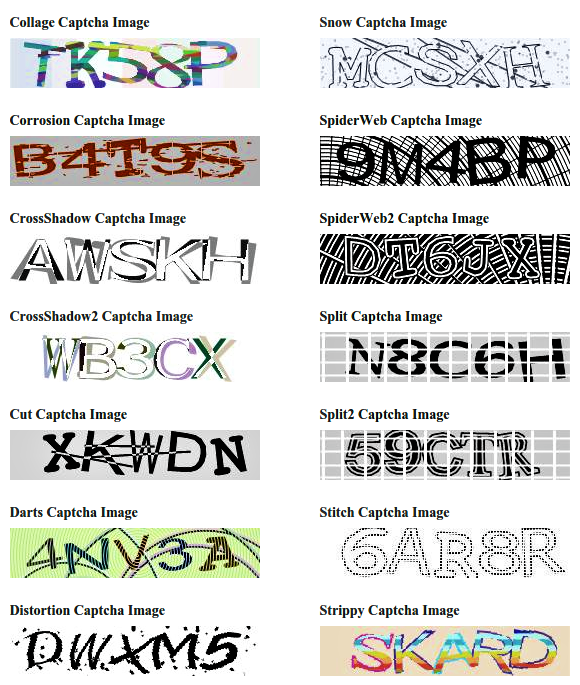}
\caption{Samples of different alphanumerical CAPTCHAs}
\label{sample-fig}
\end{figure}

In particular, these attacks often lead to situations when computer programs substitute humans, and it tries to automate services to send a considerable amount of unwanted emails, access databases, or influence the online pools or surveys \cite{p1}. One of the most common forms of cyber-attacks is the DDOS \cite{ddos} attack in which the target service is overloaded with unexpected traffic either to find the target credentials or to paralyse the system, temporarily. One of the classic yet very successful solutions is utilising a CAPTCHA system in the evolution of the cybersecurity systems. Thus, the attacking machines can be distinguished, and the unusual traffics can be banned or ignored to prevent the damage. In general, the intuition behind the CAPTCHA is
a task that can distinguish humans and machines by offering them problems that humans can quickly answer, but the machines may find them difficult, both due to computation resource requirements and the algorithm complexity \cite{ahn2013}. CAPTCHAs can be in form of numerical or alpha-numerical strings, voice, or image sets. Figure \ref{sample-fig} shows a few samples of the common alpha-numerical CAPTCHAs and their types. 

\begin{figure}[t]
\vspace{2mm}
\centering
\begin{tabular}{cccc}
    \includegraphics[width=0.35\columnwidth]{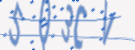} & 
    \includegraphics[width=0.35\columnwidth]{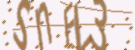} & \vspace{1mm} \\
   \end{tabular}
   \caption{Examples of a five-digit text-based CAPTCHA image.}
  \label{captcha_image}
	\vspace{-2mm}
\end{figure}

One of the commonly used practices is using text-based CAPTCHAs. An example of these types of questions can be seen in Figure \ref{captcha_image}, in which a sequence of random alphanumeric characters or digits or combinations of them are distorted and drawn in a noisy image. There are many techniques and fine-details to add efficient noise and distortions to the CAPTCHAs to make them more complex. For instance \cite{p1} and \cite{yousef} recommends several techniques to add various type of noise to improve the security of CAPTCHAs schemes such as adding crossing lines over the letters in order to imply an anti-segmentation schema. Although these lines should not be longer than the size of a letter; otherwise, they can be easily detected using a line detection algorithm. Another example would be using different font types, size, and rotation at the character level. One of the recent methods in this regard can be found in \cite{visualCryptography} which is called Visual Cryptography.

On the other hand, there are a few critical points to avoid while creating CAPTCHAs. For example, overestimating the random noises; as nowadays days the computer vision-based algorithms are more accurate and cleverer in avoiding noise in contrast to humans. Besides, it is better to avoid very similar characters such as the number '0' and the letter 'O', letter 'l' and 'I' which cannot be easily differentiated, both by the computer and a human.

\begin{figure}[t]
\centering
\includegraphics[width=0.7\columnwidth]{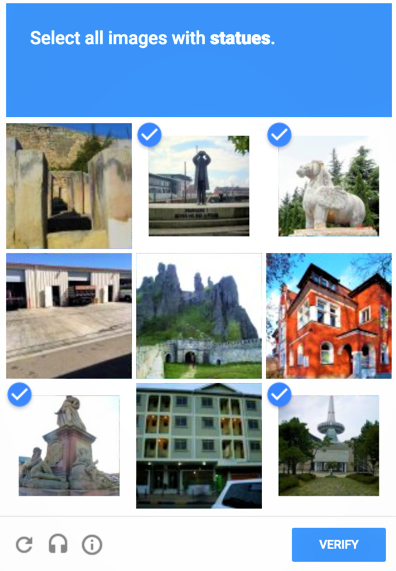}
\vspace{-1mm}
\caption{A sample of recently became accessible CAPTCHAs}
\label{image-captcha}
\end{figure}

Besides the text-based CAPTCHAs, other types of CAPTCHAs are getting popular recently. One example would be image-based CAPTCHAs that include sample images of random objects such as street signs, vehicles, statues, or landscapes and asks the user to identify a particular object among the given images \cite{imNotHuman}. These types of CAPTCHAs are especially tricky due to the context-dependent spirit. Figure \ref{image-captcha} shows a sample of this type of CAPTCHAs. 

However, in this paper, we will focus on text-based CAPTCHAs as they are more common in high traffic and dense networks and websites due to their lower computational cost. 

Before going to the next section, we would like to mention another application of the CAPTCHA systems that need to be discussed, which is its application in OCR (Optical Character Recognition) systems. Although current OCR algorithms are very robust, they still have some weaknesses in recognising different hand-written scripts or corrupted texts, limiting the usage of these algorithms. Utilising CAPTCHAs proposes an excellent enhancement to tackle such problems, as well. Since the researchers try to algorithmically solve CAPTCHA challenges this also helps to improve OCR algorithms \cite{kaurBehal}. Besides, some other researchers, such as Ahn et al. \cite{AhnRecaptcha}, suggest a systematic way to employ this method. The proposed solution is called reCAPTCHA, and it merely offers a web-based CAPTCHA system that uses the inserted text to fine-tune its OCR algorithms. The system consists of two parts: First, the preparation stage which utilises two OCR algorithms to transcribe the document independently. Then the outputs are compared, and then the matched parts are marked as correctly solved; and finally, the users choose the mismatched words to create a CAPTCHA challenge dataset \cite{multiDigit}.

This research tries to solve the CAPTCHA recognition problem, to detect its common weaknesses and vulnerabilities, and to improve the technology of generating CAPTCHAs, to ensure it will not lag behind the ever-increasing intelligence of bots and scams.

The rest of the paper is organised as follows: in Section~\ref{related}, we review on the literature by discussing the latest related works in the field. Then we introduce the details of the proposed method in Section~\ref{methodology}. The experimental results will be provided in Section~\ref{experiments}, followed by the concluding remarks in Section~\ref{conclusion}.
%
\section{Related Works}\label{related}

In this this section, we briefly explore some of the most important and the latest works done in this field. 

Geetika Garg and Chris Pollett \cite{Grag} performed a trained Python-based deep neural network to crack fix-lengthed CAPTCHAs. The network consists of two Convolutional Maxpool layers, followed by a dense layer and a Softmax output layer. The model is trained using SGD with Nesterov momentum. Also, they have tested their model using recurrent layers instead of simple dense layers. However, they proved that using dense layers has more accuracy on this problem.

In another work done by Sivakorn et al. \cite{iAmRobot}, they have created a web-browser-based system to solve image CAPTCHAs. Their system uses the Google Reverse Image Search (GRIS) and other open-source tools to annotate the images and then try to classify the annotation and find similar images, leading to an 83\% success rate on similar image CAPTCHAs. 

Stark et al. \cite{Stark} have also used a Convolutional neural network to overcome this problem. However, they have used three Convolutional layers followed by two dense layers and then the classifiers to solve six-digit CAPTCHAs. Besides, they have used a technique to reduce the size of the required training dataset.

In researches done in \cite{p1} and \cite{yousef} the authors suggest addition of different types of noise including crossing line noise or point-based scattered noise to improve the complexity and security of the CAPTCHAs patterns.

Furthermore, in \cite{breakingMicrosoft}, \cite{breakingCaptcha}, \cite{captchaBreaking}, and \cite{recognitionActive}, also CNN based methods have been proposed to crack CAPTCHA images. \cite{imageGenStyle} has used CNN via the Style Transfer method to achieve a better result. \cite{recognitionDeepCNN} has also used CNN with a small modification, in comparison with the DenseNet \cite{denseNet} structure instead of common CNNs. Also, \cite{chinese} and \cite{security} have researched Chinese CAPTCHAs and employed a CNN model to crack them. On the other hand, there are other approaches which do not use convolutional neural networks, such as \cite{optimizedSystem}. They use classical image processing methods to solve CAPTCHAs. As another example, \cite{endNigh} uses a sliding window approach to segment the characters and recognise them one by one.

Another fascinating related research field would be the adversarial CAPTCHA generation algorithm. Osadchy et al. \cite{noBotExpects} add an adversarial noise to an original image to make the basic image classifiers misclassifying them, while the image still looks the same for humans. \cite{imageGenGAN} also uses the same approach to create enhanced text-based images. 

Similarly, \cite{generativeVision} and \cite{yetAnother}, use the Generative Models and Generative Adversarial Networks from different point of views to train a better and more efficient models on the data.

\begin{figure*}[t]
\centerline{\includegraphics[width=0.85\linewidth]{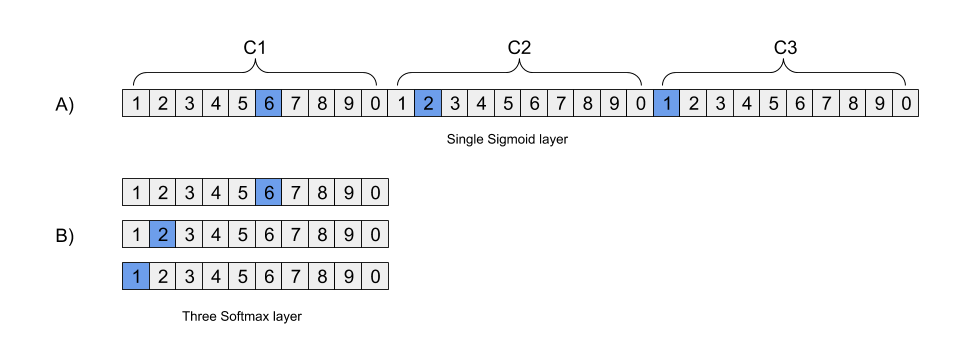}}
\vspace{-5mm}
\caption{In this figure, both approaches for representing the output data are illustrated. Method \textbf{A)} in this method, we have one single Sigmoid layer, which represents three numerical characters (3-digits CAPTCHA, each of which can represent ten different status- 0 to. Method \textbf{B)} we have three separate Softmax layers. In this example, both methods illustrate “621”.}
\label{output_data}
\end{figure*}

%
\section{Proposed Method}\label{methodology}
Deep learning based methodologies are widely used in almost all aspects of our life, from surveillance systems to autonomous vehicles \cite{autonomous}, Robotics, and even in the recent global challenge of the COVID-19 pandemic \cite{covid19}. 

To solve the CAPTCHA problem, we develop a deep neural network architecture named \textit{Deep-CAPTCHA} using customised convolutional layers to fit our requirements. Below, we describe the detailed procedure of processing, recognition, and cracking the alphanumerical CAPTCHA images. The process includes input data pre-processing, encoding of the output, and the network structure itself.
%
\subsection{Preprocessing}\label{processing}
Applying some pre-processing operations such as image size reduction, colour space conversion, and noise reduction filtering can have a tremendous overall increase on the network performance. 

The original size of the image data used in this research is $135 \times 50$ pixel which is too broad as there exist many blank areas in the CAPTCHA image as well as many co-dependant neighbouring pixels. Our study shows by reducing the image size down to $67 \times 25$ pixel, we can achieve almost the same results without any noticeable decrease in the system’s performance. This size reduction can help the training process to become faster since it reduces the data without having much reduction in the data entropy.

Colour space to Gray-space conversion is another preprocessing method that we used to reduce the size of the data while maintaining the same level of detection accuracy. In this way, we could further reduce the amount of redundant data and ease the training and prediction process. Converting from a three-channel RGB image to a grey-scale image does not affect the results, as the colour is not crucial on the text-based CAPTCHA systems.

The last preprocessing technique that we consider is the application of a noise reduction algorithm. After a careful experimental  analysis on the appropriate filtering approaches, we decided to implement the conventional Median-Filter to remove the noise of the input image. The algorithm eliminates the noise of the image by using the median value of the surrounding pixels values instead of the pixel itself. The algorithm is described in \textit{Algorithm} \ref{alg_median} in which we generate the \textit{resultImage} from the input \textit{'image'} using a predefined \textit{window size}.
\begin{algorithm}[t]
\SetAlgoLined
\SetKwInput{KwInput}{Input}
\SetKwInput{KwOutput}{Output}
\KwInput{image, window size}
\KwOutput{resultImage}
\For{x from 0 to image width}
{
  \For{y from 0 to image height}
   {
    i = 0
    \For{fx from 0 to window width}{
        \For{fy from 0 to window height}{
            window[i] := image[x + fx][y + fy];\\
            i := i + 1;
        }
    }
    sort entries in window; \\
    resultImage[x][y] := window[window width * window height / 2];
  }
}
\caption{Median filter noise filtering}
\label{alg_median}
\vspace{-2mm}
\end{algorithm}

\subsection{Encoding}\label{encoding}
Unlike the classification problems where we have a specific number of classes in the CAPTCHA recognition problems, the number of classes depends on the number of digits and the length of the character set in the designed CAPTCHA. This leads to exponential growth depending on the number of classes to be detected. Hence, for a CAPTCHA problem with five numerical digits, we have around 100,000 different combinations. As a result, we are required to encode the output data to fit into a single neural network.

The initial encoding we used in this research was to employ $nb\_input = D \times L$ neurons, where $D$ is the length of the alphabet set, and $L$ is the character set length of the CAPTCHA. The layer utilises the Sigmoid activation function:

\begin{equation}
\label{eq_sigmoid}
S(x)=\frac {1}{1+e^{-x}}
\end{equation}

Where $x$ is the input value and $S(x)$ is the output of the Sigmoid function. By increasing the $x$, the $S(x)$ conversing to $1$ and by reducing it the $S(x)$ is getting close to $-1$. Applying Sigmoid function adds a non-linearity feature to neurons which improves the learning potential and also the complexity of those neurons in dealing with non-linear inputs. 

These sets of neurons can be arranged in a way so that the first set of $D$ neurons represent the first letter of the CAPTCHA; the second set of $D$ neurons represent the second letter of the CAPTCHA, and so on. In other words, assuming $D = 10$, the $15^{th}$ neuron tells whether the fifth letter from the second character matches with the predicted alphabet or not. A visual representation can be seen in Figure \ref{output_data}.A, where the method encompasses three numerical serial digits that represent 621 as the output.
However, this approach seemed not to be worthy due to its incapability of normalising the numerical values and also the impossibility of using the Softmax function as the output layer of the intended neural network.

Therefore, we employed $L$ parallel Softmax layers, instead:

\begin{equation}
\label{eq_softmax}
\sigma ( {z} )_i=\frac {e^{z_i}}{\sum _{j=1}^{K}e^{z_j}}
\end{equation}

where $i$ is the corresponding class for which the Softmax is been calculated, $z_i$ is the input value of that class, and $K$ is the maximum number of classes.

Each Softmax layer individually represents $D$ neurons as Figure \ref{output_data}.B and these $D$ neurons in return represent the alphabet that is used to create the CAPTCHAs (for example 0 to 9, or A to Z). 

$L$ unit is represents the location of the digit in the CAPTCHA pattern (for example, locations 1 to 3). Using this technique allows us to normalise each Softmax unit individually over its neurons, $D$. In other words, each unit can normalise its weight over the different alphabets; hence it performs better, in overall.

\begin{figure*}[t]
\centerline{\includegraphics[width=1.0\textwidth]{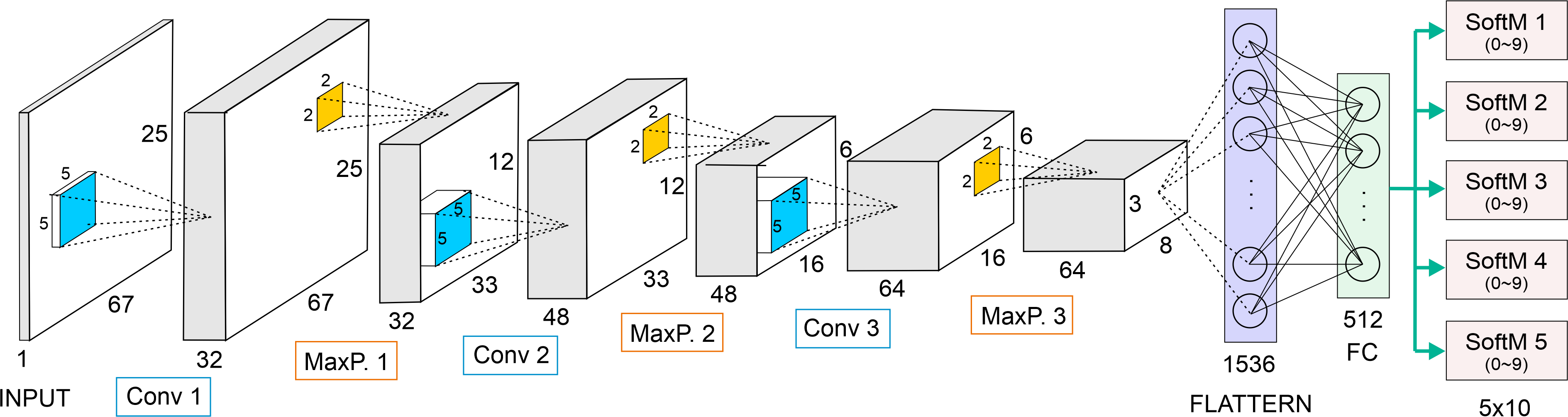}}
\caption{The Architecture of the proposed \textit{Deep-CAPTCHA} Network}
\label{Architecture}
\end{figure*}

%
\subsection{Architecture of the Network}\label{architecture}
Although the Recurrent Neural Networks (RNNs) can be one of the options to predict CAPTCHA characters, in this research we have focused on sequential models as they perform faster than RNNs, yet can achieve very accurate results if the model is well designed.

The structure of our proposed network is depicted in Figure \ref{Architecture}. The network starts with a Convolutional layer with 32 input neurons, the ReLU activation function, and $5 \times 5$ Kernels. A $2\times 2$ Max-Pooling layer follows this layer. Then, we have two sets of these Convolutional-MaxPooling pairs with the same parameters except for the number of neurons, which are set to 48 and 64, respectively. We have to note that all of the Convolutional layers have the $``same"$ padding parameter.

After the Convolutional layers, there is a 512 dense layer with the ReLU activation function and a 30\% drop-out rate. Finally, we have $L$ separate Softmax layers, where $L$ is the number of expected characters in the CAPTCHA image.

\begin{figure*}[hbt!]
  \centering
  \begin{subfigure}[b]{0.475\linewidth}
     \includegraphics[scale=0.17]{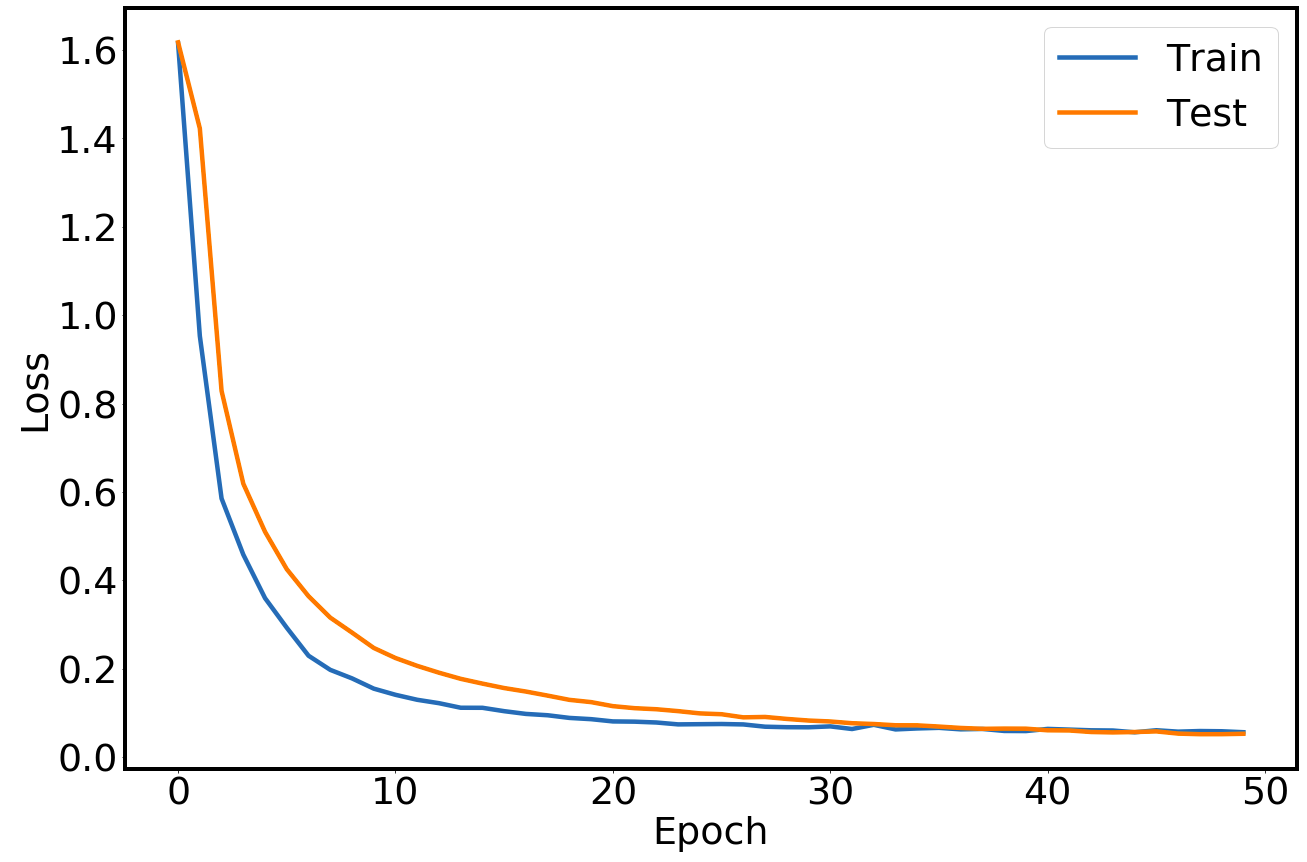}
    \caption{Loss - Adam}
    \label{loss_adam}
  \end{subfigure}	
  \begin{subfigure}[b]{0.475\linewidth}
    \includegraphics[scale=0.17]{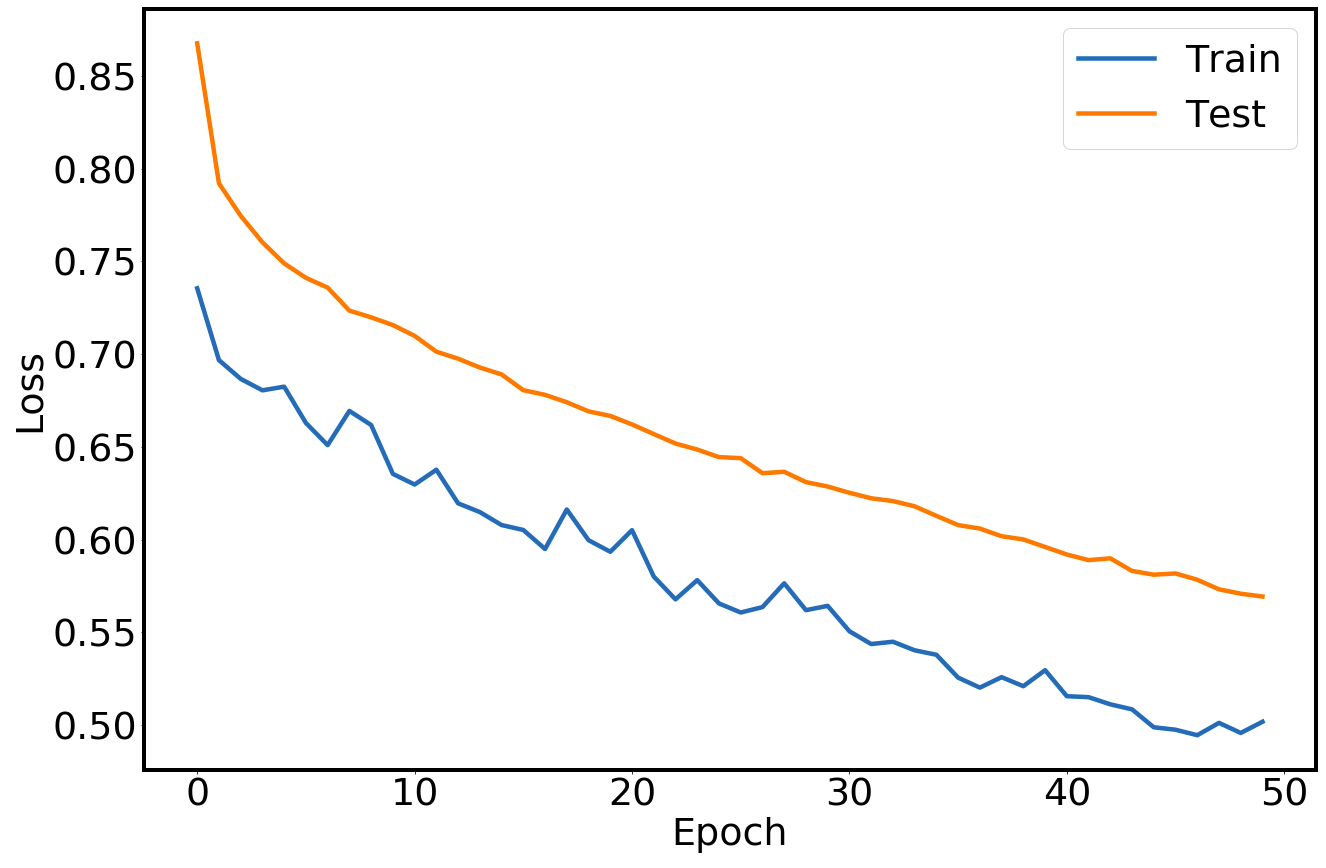}
    \caption{Loss - SGD}
    \label{loss_sgd}
  \end{subfigure}
	  \begin{subfigure}[b]{0.475\linewidth}
    \includegraphics[scale=0.17]{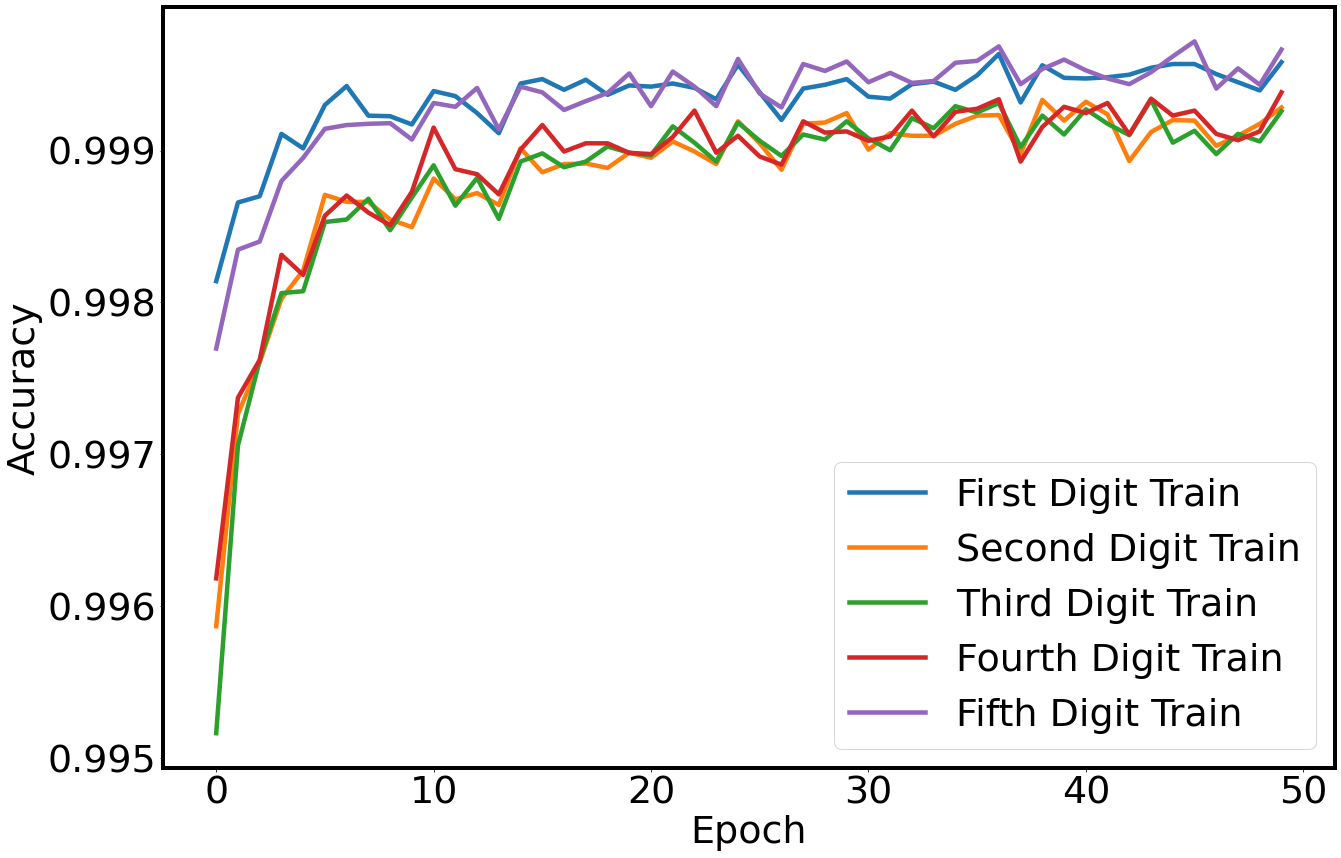}
    \caption{Train Accuracy - Adam}
    \label{acc_train_adam}
  \end{subfigure}
	  \begin{subfigure}[b]{0.475\linewidth}
    \includegraphics[scale=0.17]{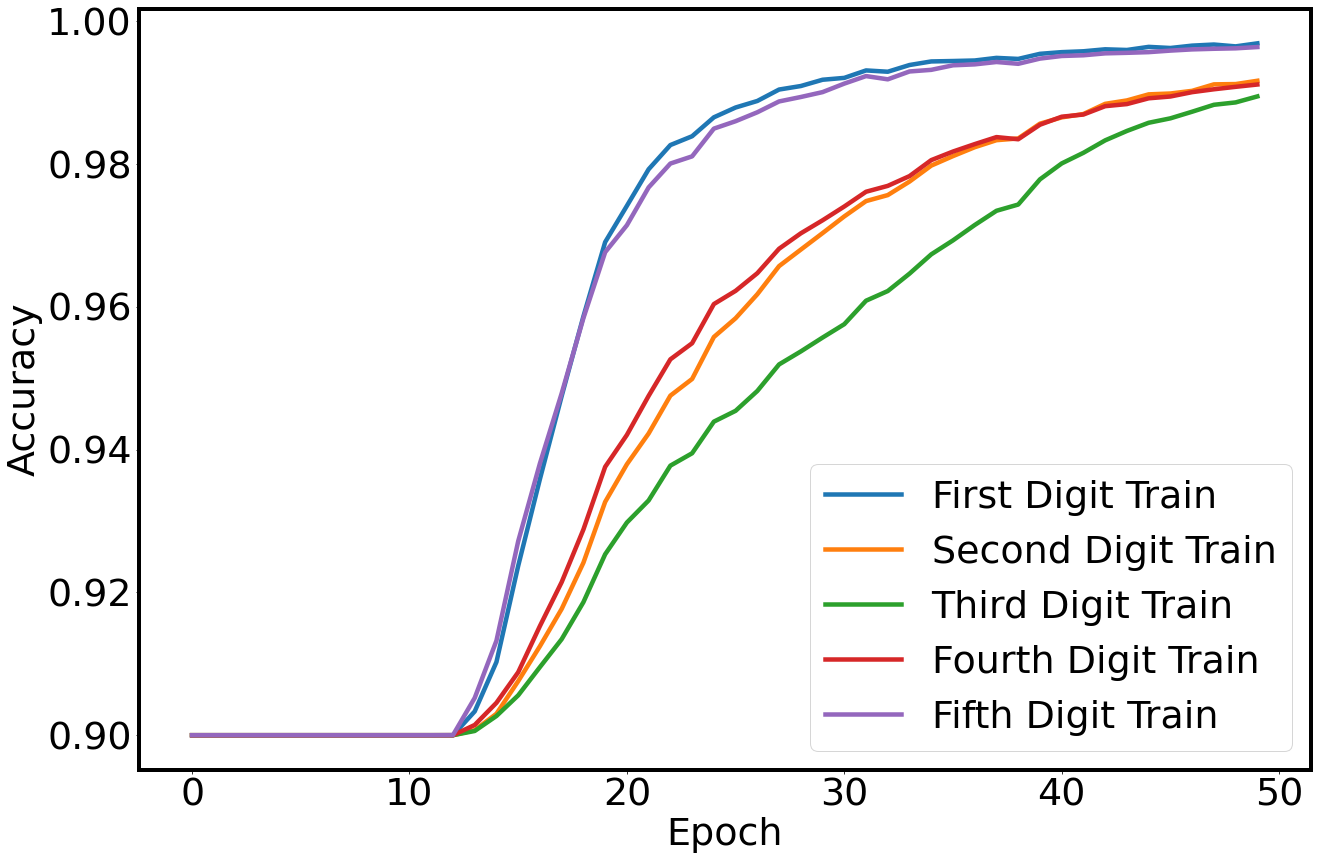}
    \caption{Train Accuracy - SGD}
    \label{acc_train_sgd}
  \end{subfigure}
	  \begin{subfigure}[b]{0.485\linewidth}
    \includegraphics[scale=0.17]{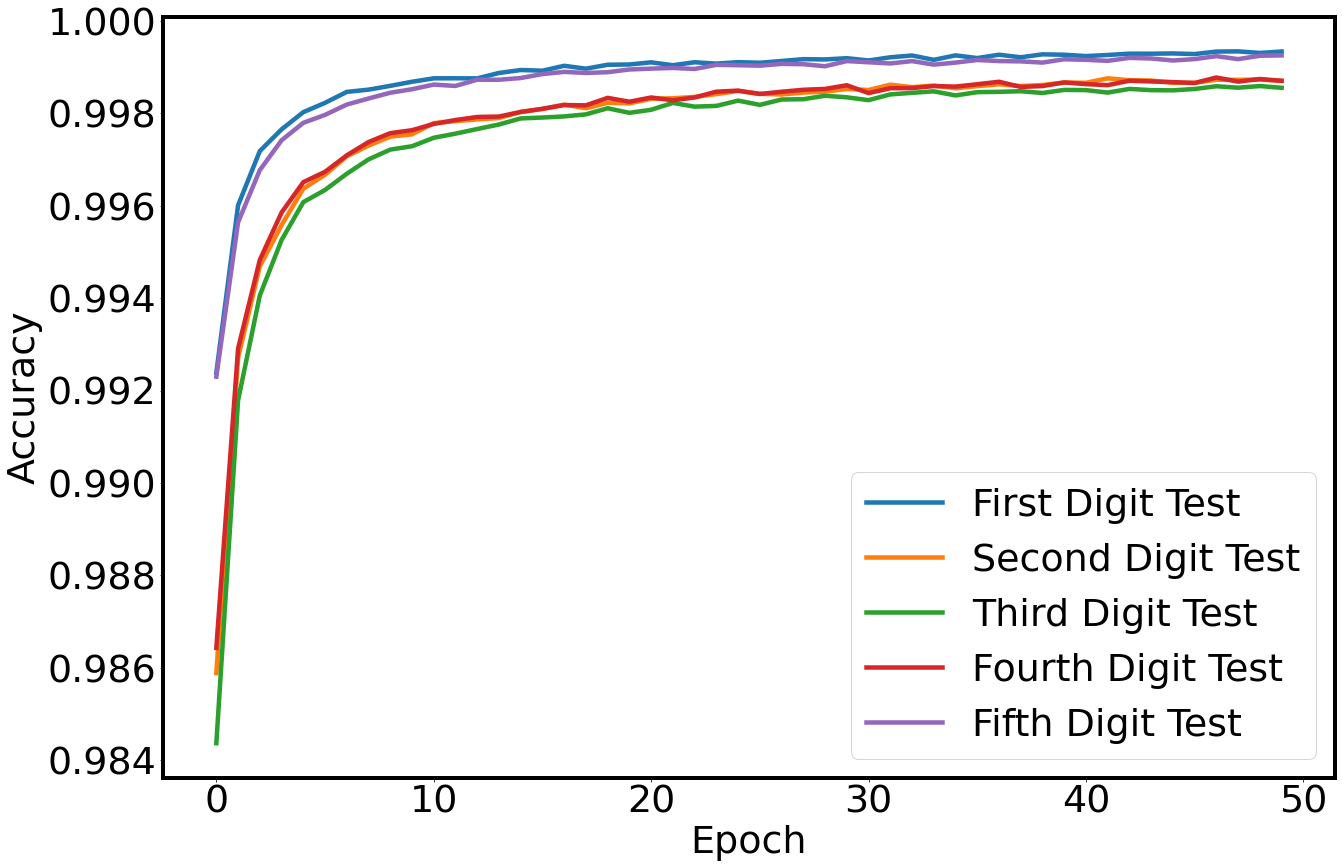}
    \caption{Test Accuracy - Adam}
    \label{acc_test_adam}
  \end{subfigure}
  \begin{subfigure}[b]{0.485\linewidth}
    \includegraphics[scale=0.17]{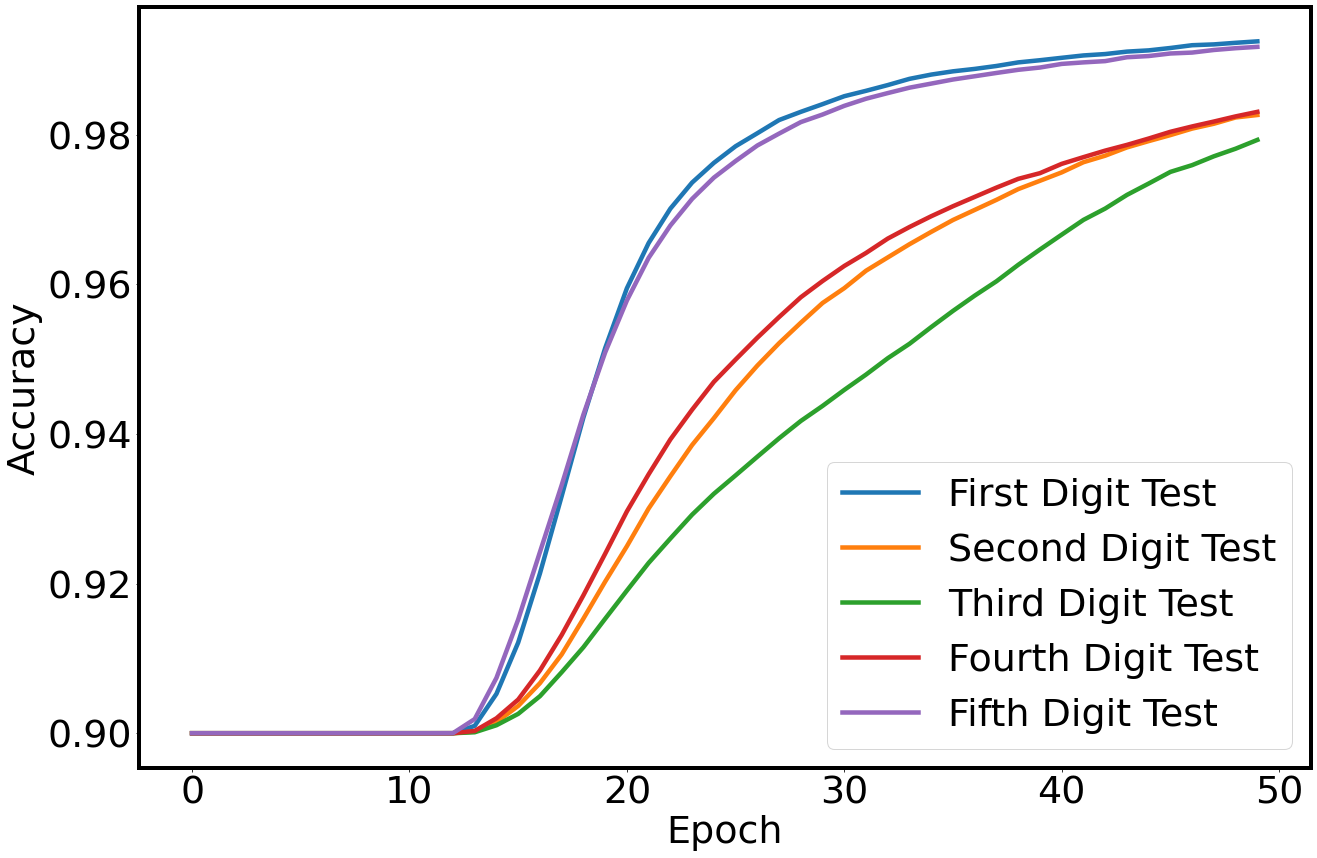}
    \caption{Test Accuracy - SGD}
    \label{acc_test_sgd}
  \end{subfigure}
   \caption{(a) and (b): Loss values of the test and training  process for Adam and SGD optimiser, respectively. (c) and (d): The accuracy metrics of the network for the same optimisers on the training dataset. (e) and (f): The accuracy metrics of the network using the given optimisers on the test dataset.}
  \label{fig_learning_curve}
\end{figure*}

The loss function of the proposed network is the Binary-cross entropy as we need to compare these binary matrices all together:

\begin{equation}
\label{eq_binary_cross}
H_p(q) = -\frac{1}{N} \sum_{i=1}^{N} y_i.\log(p(x_i))+(1-y_i).\log(1-p(x_i))
\end{equation}

were $N$ is the number of samples and $p$ is the predictor model. The $x_i$ and $y_i$ represent the input data and the label of the $i^{th}$ sample, respectively. Since the label could be either zero or one, therefore only one part of this equation would be active for each sample.

We also employed Adam optimiser, which is briefly described in Equations \ref{eq_adam_1} to \ref{eq_adam_5} where $m_t$ and $v_t$ representing an exponentially decaying average of the past gradients and past squared gradients, respectively. 

\begin{equation}
\label{eq_adam_1}
m_t = \beta_1 m_{t-1} + (1 - \beta_1) g_t
\end{equation}
\begin{equation}
\label{eq_adam_2}
v_t = \beta_2 v_{t-1} + (1 - \beta_2) g_t^2
\end{equation}

$\beta_1$ and $\beta_2$ are configurable constants. $g_t$ is the gradient of the optimising function and $t$ is the learning iteration. In Equations \ref{eq_adam_3} and \ref{eq_adam_4}, momentary values for $m$ and $v$ are calculated as follows:

\begin{equation}
\label{eq_adam_3}
\hat{m}_t = \frac{m_t}{1 - \beta_1^t}
\end{equation}

\begin{equation}
\label{eq_adam_4}
\hat{v}_t = \frac{v_t}{1 - \beta_2^t}
\end{equation}

Finally, using Equation \ref{eq_adam_5} and by updating $\theta_t$ in each iteration, the optimum value of the function could be attained. $\hat{m}_t$ and $\hat{v}_t$ are calculated via Equations \ref{eq_adam_3} and \ref{eq_adam_4} and $\eta$, the step size (also known as learning rate) is set to 0.0001 in our approach.

\begin{equation}
\label{eq_adam_5}
\theta_{t+1} = \theta_t - \frac{\eta}{\sqrt{\hat{v}_t} + \epsilon} \hat{m}_t
\end{equation}

The intuition behind using Adam optimiser is its capability in training the network in a reasonable time. This can be easily inferred from Figure \ref{loss_adam} in which the Adam optimiser achieves the same results in comparison with Stochastic Gradient Descent (SGD), but with a much faster convergence. 

After several experiments, we trained the network for 50 epochs with a batch size of 128 for each. As can be inferred from Figure \ref{loss_adam}, even after 30 epochs the network tends to an acceptable convergence. As a result, 50 epochs seem to be sufficient for the network to perform steadily. Furthermore, Figure \ref{acc_test_adam} would also suggest the same inference based on the measured accuracy metrics.

\section{Experimental Results}\label{experiments}
\begin{figure}[t]
\centering
\begin{tabular}{cccc}
    \includegraphics[width=0.3\columnwidth]{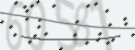} & 
    \includegraphics[width=0.3\columnwidth]{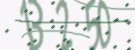} & \vspace{1mm} \\
    \includegraphics[width=0.3\columnwidth]{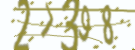} & 
    \includegraphics[width=0.3\columnwidth]{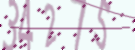} & \vspace{1mm} \\
    \includegraphics[width=0.3\columnwidth]{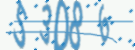} & 
    \includegraphics[width=0.3\columnwidth]{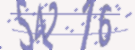} & \vspace{1mm} \\
   \end{tabular}
   \caption{Samples of the Python numerical Image-Captcha Library used to train the \textit{Deep-CAPTCHA}.}
  \label{fig_six_captcha}
\end{figure}

After developing the above-described model, we trained the network on 500,000 randomly generated CAPTCHAs using Python ImageCaptcha Library \cite{pythonCaptchaLib}. See Figure \ref{fig_six_captcha} for some of the randomly generated numerical CAPTCHAs with the fixed lengths of five-digits. 

\begin{figure*}[t]
\centerline{\includegraphics[width=0.9\textwidth]{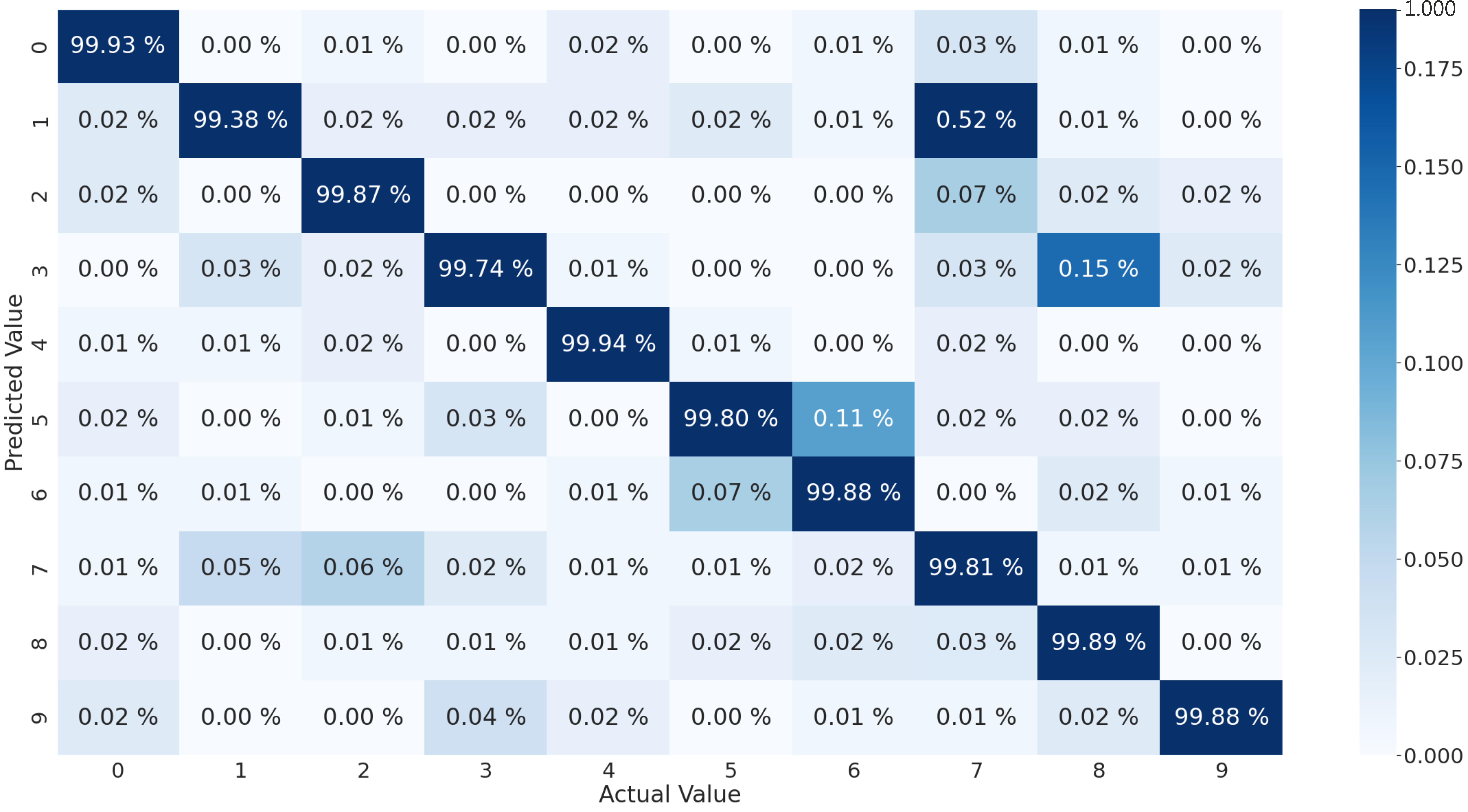}}
\caption{Confusion Matrix of the trained network on the test data. Although the digits are accurately labelled, the vertical bar is scaled non-linearly to point out the differences with a better visualisation.}
\label{confusion_test}
\end{figure*}

To be balanced, the dataset consists of ten randomly generated images from each permutation of a five-digit text.

\begin{table}[t]
\caption{Accuracy metric and total loss value for the Train and Test portion of the dataset.}
\label{table_result}
\begin{center}
\begin{tabular}{c|c|c|c|c}
\hline
      & \multicolumn{2}{c|}{\rule{0pt}{3ex} \textbf{Softmax}} & \multicolumn{2}{c}{\textbf{Sigmoid}}  \\ 
			[+0.5em]
\cline{2-5}
      & \textbf{\rule{0pt}{3ex}Loss} & \textbf{Accuracy} & \textbf{Loss} & \textbf{Accuracy} \\ 
			[+0.5em]
\hline
\textbf{\rule{0pt}{3ex}Train} & 0.013 & 99.33\%              & 0.037   & 98.73\%                     \\
\textbf{\rule{0pt}{3ex}Test}  & 0.075 & 98.94\%              & 0.116   & 90.04\%                    
\end{tabular}
\end{center}
\end{table}

\begin{table}[t]
\caption{Accuracy metric of the dataset for each digit and the complete CAPTCHA as a set of 5 integrated digits.}
\label{table_result_digits}
\begin{center}
\begin{tabular}{c|c|c}
\hline
\textbf{} & \textbf{\rule{0pt}{3ex} Train Accuracy} & \textbf{Test Accuracy} \\ 
[+0.3em]
\hline
\rule{0pt}{3ex}\textbf{Digit 1}     & 99.91\%     & 99.87\%       \\ 
[+0.3em]
\textbf{Digit 2}     & 99.85\%     & 99.75\%       \\ 
[+0.3em]
\textbf{Digit 3}     & 99.84\%     & 99.72\%       \\ 
[+0.3em]
\textbf{Digit 4}     & 99.83\%     & 99.75\%       \\ 
[+0.3em]
\textbf{Digit 5}     & 99.90\%     & 99.85\%       \\ 
[+0.3em]
\textbf{CAPTCHA}     & 99.33\%     & 98.94\%       \\ 
\end{tabular}
\end{center}
\end{table}

\subsection{Performance Analysis}
We tested the proposed model on another set of half a million CAPTCHA images as our test dataset. As represented in Table \ref{table_result}, the network reached the overall performance and accuracy rate of 99.33\% on the training set and 98.94\% on the test dataset. We have to note that the provided accuracy metrics are calculated based on the number of correctly detected CAPTCHAs as a whole (i.e. correct detection of all five individual digits in a given CAPTCHA); otherwise, the accuracy of individual digits are even higher, as per the Table \ref{table_result_digits}.

We have also conducted a confusion matrix check to visualise the outcome of this research better.  Figure \ref{confusion_test} shows how the network performs on each digit regardless of the position of that digit in the CAPTCHA string. As a result, the network seems to work extremely accurately on the digits, with less than 1\% misclassification for each digit.

\subsection{Vulnerability Assessment}

By analysing the network performance and visually inspecting 100 misclassified samples we pointed out some important results as follows that can be taken into account to decrease the vulnerability of the CAPTCHA generators: \\

While an average human could solve the majority of the misclassified CAPTCHAs, the following weaknesses were identified in our model that caused failure by the \textit{Deep-CAPTCHA} solver:
\begin{itemize}
\item  In 85\% of the misclassified samples, the gray-level intensity of the generated CAPTCHAs were considerably lower than the average intensity of the Gaussian distributed pepper noise in the CAPTCHA Image. 
\item In 54\% of the cases, the digits 3, 8, or 9 were the cause of the misclassification. 
\item In 81.8\% of the cases, the misclassified digits were rotated for $10^\circ$ or more.
\item Confusion between the digits 1 and 7 was also another cause of the failures, particularly in case of more than $20^\circ$ counter-clockwise rotation for the digit 7.
\end{itemize}
Consequently, in order to cope with the existing weakness and vulnerabilities of the CAPTCHA generators, we strongly suggest mandatory inclusion of one or some of the digits 3, 7, 8 and 9 (with/without counter-clockwise rotations) with a significantly higher rate of embedding in the generated CAPTCHAs comparing to the other digits. 
This will make the CAPTCHAs harder to distinguish for automated algorithms such as the \textit{Deep-CAPTCHA}, as they are more likely to be confused with other digits, while the human brain has no difficulties in identifying them.

A similar investigation was conducted for the alphabetic part of the failed detections by the \textit{Deep-CAPTCHA} and the majority of the unsuccessful cases were tied to 
either too oriented characters or those with close contact to neighbouring characters.
For instance, the letter $``g"$ could be confused with $``8"$ in certain angles, or a $``w"$ could be misclassified as an $``m"$ while contacting with an upright letter such as $``T"$. In general, the letters that can tie together with one/some of the letters: w, v, m, n can make a complex scenario for the \textit{Deep-CAPTCHA}. Therefore we suggest more inclusion of these letters, as well as putting these letters in close proximity to others letter, may enhance the robustness of the CAPTCHAs.


Our research also suggests brighter colour (i.e. lower grayscale intensity) alpha-numerical characters would also help to enhance the difficulty level of the CAPTCHAs.
%
\subsection{Performance Comparison}
In this section, we compare the performance of our proposed method with 10 other state-of-the-art techniques. The comparison results are illustrated in Table \ref{table_result_compared} followed by further discussions about specification of each method. 

As mentioned in earlier sections, our approach is based on Convolutional Neural Network that has three pairs of Convolutional-MaxPool layers followed by a dense layer that is connected to a set of Softmax layers. Finally, the network is trained with Adam optimiser. 

In this research we initially focused on optimising our network to solve numerical CAPTCHAs; However, since many existing methods work on both numerical and alphanumerical CAPTCHAs, we developed another network capable of solving both types. Also, we trained the network on 700,000 alphanumerical CAPTCHAs. For a better comparison and to have a more consistent approach, we only increased the number of neurons in each Softmax units from 10 to 31 to cover all common Latin characters and digits.

The reason behind having 31 neurons is that we have used all Latin alphabets and numbers except for ${i, l, 1, o, 0}$ due to their similarity to each other and existing difficulties for an average human to tell them apart. Although we have used both upper and lower case of each letter to generate a CAPTCHA, we only designate a single neuron for each of these cases in order to simplicity.


In order to compare our solution, first, we investigated the research done by Wang et al. \cite{recognitionDeepCNN} which includes evaluations on the following approaches: DenseNet-121 and ResNet-50 which are fine-tuned model of the original DenseNet and ResNet networks to solve CAPTCHAs as well as DFCR which is an optimised method based on the DenseNet network. The DFCR has claimed an accuracy of 99.96\% which is the best accuracy benchmark among other methods. However, this model has only been trained on less than 10,000 samples and only on four-digit CAPTCHA images. Although the quantitative comparison in Table \ref{table_result_compared} shows the \cite{recognitionDeepCNN} on top of our proposed method, the validity of the method can neither be verified on larger datasets, nor on complex alphanumerical CAPTCHAs with more than half a million samples, as we conducted in our performance evaluations.

The next comparing method is \cite{svm} which uses an SVM based method and also  implementation of the VGG-16 network to solve CAPTCHA problems. The critical point of this method is the usage of image preprocessing, image segmentation and one by one character recognition. These techniques have lead to 98.81\% accuracy on four-digit alphanumerical CAPTCHAs. The network has been  trained on a dataset composed of around 10,000 images. Similarly, TOD-CNN \cite{lowCostApproach} have utilised segmentation method to locate the characters in addition to using a CNN model which is trained on a 60,000 dataset. The method uses a TensorFlow Object Detection (TOD) technique to segment the image and characters.

\begin{table}[t]
\caption{The accuracy results of different CAPTCHA recognition methods. ($*$ denotes the accuracy result that only calculated for numerical CAPTCHAs.)}
\label{table_result_compared}
\begin{center}
\begin{tabular}{l|l|l}
\hline
\textbf{\small{References}}  & \textbf{\small{Methodology}}              & \textbf{\small{Accuracy}} \\ [0.1em] \hline
\multirow{3}{*}{Wang et al. \cite{recognitionDeepCNN}} & \rule{0pt}{2.5ex}1- DFCR & 99.96\%           \\ 
                             & \rule{0pt}{2.5ex}2- DenseNet-121     & 99.90\%                           \\ 
                             & \rule{0pt}{2.5ex}3- ResNet-50                & 99.90\%                   \\ 
														\hline
\multirow{2}{*}{\textbf{Current paper}} & \multirow{2}{9em}{\rule{0pt}{2.5ex} \textbf{Proposed~Method$^*$} (~Numerical~)}  & 
														\multirow{2}{*}{\rule{0pt}{2.5ex} \textbf{98.90\%}} \\ 
												 		& \rule{0pt}{2.5ex}                    &                   \\ 
														\hline
\multirow{2}{*}{Wei et al. \cite{svm}} & \rule{0pt}{2.5ex}4- SVM          & 98.81\%                   \\ 
                             & \rule{0pt}{2.5ex}5- CNN                      & 98.43\%                   \\ 
														\hline
\multirow{2}{*}{\textbf{Current paper}} & \multirow{2}{9em}{\rule{0pt}{2.5ex} \textbf{Proposed~Method} (~Alphanumerical~)} & 
														\multirow{2}{*}{\rule{0pt}{2.5ex} \textbf{98.30\%}}  \\ 
														& \rule{0pt}{2.5ex}                    &                   \\ \hline
Goodfellow \cite{multiDigit}            & \rule{0pt}{2.5ex}6- SVHN Network $^*$        & 97.84\%                   \\ \hline
\multirow{3}{*}{Du et al. \cite{r-cnn}}   & \rule{0pt}{2.5ex}7- VGG 16    & 97.50\%                   \\ 
                             & \rule{0pt}{2.5ex}8- VGG\_CNN\_M\_1024        & 97.20\%                   \\ 
                             & \rule{0pt}{2.5ex}9- ZF                       & 96.60\%                   \\ \hline
Yu et al. \cite{lowCostApproach}       & \rule{0pt}{2.5ex}10-TOD-CNN                  & 92.37\%                   \\ \hline
\end{tabular}
\end{center}
\end{table}

Goodfellow et al. \cite{multiDigit} have used DistBelief implementation of CNNs to recognise numbers more accurately. The dataset used in this research was the Street View House Numbers (SVHN) which contains images taken from Google Street View. 

Finally, the last discussed approach is \cite{r-cnn} which compares VGG16, VGG\_CNN\_M\_1024, and ZF. Although they have relatively low accuracy compared to other methods, they have employed R-CNN methods to recognise each character and locate its position at the same time. 

In conclusion, our methods seem to have relatively satisfactory results on both numerical and alphanumerical CAPTCHAs. Having a simple network architecture allows us to utilise this network for other purposes with more ease. Besides, having an automated CAPTCHA generation technique allowed us to train our network with a better accuracy while maintaining the detection of more complex and more comprehensive CAPTCHAs comparing to state-of-the-art.

\section{Conclusion}\label{conclusion}
We designed, customised and tuned a CNN based deep neural network for numerical and alphanumerical based CAPTCHA detection to reveal the strengths and weaknesses of the common CAPTCHA generators. Using a series of paralleled Softmax layers played an important role in detection improvement. We achieved up to 98.94\% accuracy in comparison to the previous 90.04\% accuracy rate in the same network, only with Sigmoid layer, as described in section \ref{encoding} and Table \ref{table_result}.

Although the algorithm was very accurate in fairly random CAPTCHAs, some particular scenarios made it extremely challenging for \textit{Deep-CAPTCHA} to crack them.  We believe taking the addressed issues into account can help to create more reliable and robust CAPTCHA samples which makes it more complex and less likely to be cracked by bots or AI-based cracking engines and algorithms.

As a potential pathway for future works, we suggest solving the CAPTCHAs with variable character length, not only limited to numerical characters but also applicable to combined challenging alpha-numerical characters as discussed in section \ref{experiments}. We also recommend further research on the application of Recurrent Neural Networks as well as the classical image processing methodologies \cite{objectdetection} to extract and identify the CAPTCHA characters, individually.

\end{document}